%% file: main.tex
\documentclass[11pt]{article}

\PassOptionsToPackage{table}{xcolor}
\usepackage[final]{acl}

\usepackage{times}
\usepackage{latexsym}
\usepackage[T1]{fontenc}
\usepackage[utf8]{inputenc}
\usepackage{microtype}
\usepackage{CJKutf8}
\usepackage{natbib}
\usepackage{url}

\usepackage{amsmath,amssymb}
\usepackage{graphicx}
\usepackage{enumitem}
\usepackage{booktabs}
\usepackage{tabularx}
\usepackage{xcolor}
\usepackage{algorithm}
\usepackage{algorithmic}
\usepackage{comment}

\usepackage[most]{tcolorbox}  % tcolorbox 功能很全
\usepackage[capitalise]{cleveref}

\newtcolorbox{promptbox}[1]{%
  enhanced,
  breakable,                 % 内容可分页
  colback=gray!5,            % 内部浅灰背景
  colframe=black,            % 外边框颜色
  boxrule=0.8pt,             % 外边框粗细
  arc=2mm,                   % 圆角（如果想完全直角，改成 0mm）
  left=2mm,right=2mm,top=2mm,bottom=2mm, % 内边距
  title={#1},                % 标题内容
  fonttitle=\bfseries,       % 标题加粗
  coltitle=white,            % 标题文字白色
  colbacktitle=black         % 标题区域黑底
}

\title{Profile-Driven User Simulation via Reinforcement Learning with Hierarchical Rewards for
Task-Oriented Dialogues
}

\author{
Juan Li\quad
Fangshu Chang \quad
Xinyi Jiang \quad
Ling Huang \quad
Yunzhang Chen\\
Bairong Inc.
}

\begin{document}
\maketitle

\date{October 15, 2026}

\input{abstract}

\input{introduction}

\input{related_work}

\input{methodology}

\input{experiments}

\input{conclusion}

\input{limitations}

\input{citations}

\clearpage
\appendix

\input{appendix}

\end{document}

%% file: abstract.tex
%% ============================================================
%%  ABSTRACT
%% ============================================================
\begin{abstract}
Building Task-Oriented Dialogue (TOD) systems for high-value scenarios such as financial marketing 
%%outbound calls 
requires large volumes of domain-specific dialogue data, which privacy regulations, annotation costs, and the need for expert labor make hard to obtain. User simulation offers a remedy, but existing methods target general-domain dialogue and exhibit two key limitations: insufficient modeling of domain-specific business processes and compliance constraints, and inadequate capture of the colloquial, spoken-language character of voice interactions. We propose \textsc{ProfileSim}, a profile-driven user simulation framework for financial marketing outbound calls. It jointly models \emph{user profiles}---demographics and Big Five personality traits---and \emph{business profiles}---intents reverse-engineered from real call logs---enabling fine-grained control over simulated behavior. The base model is fine-tuned on real voice dialogue data and refined by Group Relative Policy Optimization (GRPO) with hierarchical rewards at the dialogue and utterance levels. Paired with a customer service agent integrating Standard Operating Procedure (SOP) constraints and Retrieval-Augmented Generation (RAG), \textsc{ProfileSim} achieves the highest colloquial naturalness, 
%(4.97 vs.\ at most 4.21 for zero-shot LLM baselines), 
the strongest user profile adherence 
%(0.926 vs.\ at most 0.841), 
and the lowest agent score on long-tail intents among all evaluated user generators, while producing substantially shorter user utterances at comparable business compliance. The generated dialogue data is designed to support intent recognition training, agent evaluation, and rapid cold-start deployment.
\end{abstract}

%\noindent\textbf{Keywords:} User Simulation, Sales Outbound Calling, Profile-Driven %Dialogue Generation, Supervised Fine-Tuning, Group Relative Policy Optimization, %Retrieval-Augmented Generation

%% file: introduction.tex
%% ============================================================
%%  SECTION 1: INTRODUCTION
%% ============================================================
\section{Introduction}
\label{sec:intro}

Task-oriented dialogue (TOD) systems are central to intelligent customer service and online marketing, typically using a modular pipeline of natural language understanding, dialogue state tracking, policy learning, and natural language generation~\citep{young2013pomdp,chen2017survey,gao2019neural,zhang2020survey_dialogue,qin2023etod_survey}. In high-stakes domains such as financial services, these systems must additionally satisfy stringent compliance, accuracy, and professional-rigor requirements~\citep{wu2023bloomberggpt,yang2023fingpt}: responses about interest rates, investment returns, or risk disclosures must be grounded in structured knowledge and regulatory guidelines. Adapting general-purpose dialogue systems to such standards remains challenging because they optimize for fluency rather than domain knowledge and compliance, and high-quality labeled dialogue data is scarce due to expert annotation, regulatory review, and privacy considerations~\citep{altinok2023survey_data_scarcity}.
%~\citep{schatzmann2006survey,altinok2023survey_data_scarcity}.

User simulation---constructing virtual users that interact with dialogue systems to generate synthetic data---has progressed from agenda-based simulators~\citep{schatzmann2009hidden,li2016user_simulator} through neural approaches~\citep{asri2016sequence,kreyssig2018neural,lin2022gentus} to LLM-based simulators~\citep{davidson2023user_sim_llm,sekulic2024daus,luo2024duetsim,wang2025user_behavior_sim,ren2024bases,long2024llms_synth_survey,harsha2025synthetic_fin_qa}. Two deficiencies persist in vertical domains: (1) \textbf{insufficient domain constraints}---methods trained on general-domain corpora such as MultiWOZ~\citep{budzianowski2018multiwoz} lack effective modeling of business processes and compliance rules; (2) \textbf{inadequate spoken-language modeling}---written-text training produces outputs that diverge from the disfluencies and informality of real voice interactions~\citep{ji2024wavchat,shamardina2025llm_spoken_conv,shim2026spokenus}.

We propose \textsc{ProfileSim}, a user simulation framework for financial marketing outbound calls. The core is a \emph{joint modeling mechanism of business and user profiles}: business processes (product attributes, sales workflow, compliance) and user attributes (demographics, Big Five traits) act as explicit conditioning variables. Training on real voice dialogue data mitigates the rigidity of prompt-based generation with closed-source LLMs, producing dialogues intended to support intent recognition training, agent evaluation, and cold-start deployment.

\noindent\textbf{Contributions.} (1) A domain-specific user simulation framework that jointly models business and user profiles for financial outbound calling. (2) A training recipe that combines supervised fine-tuning (SFT) with GRPO-based preference optimization under hierarchical dialogue- and utterance-level rewards.
% improving colloquial naturalness, dialogue coherence, and persona consistency in multi-turn interactions~\citep{frisch2025consistent_persona_rl}. 
(3) Experiments on real outbound-call data showing that the method outperforms strong zero-shot LLM baselines on colloquial naturalness, user profile adherence, and downstream agent stress, and that each component contributes measurable gains.

%% file: related_work.tex
%% ============================================================
%%  SECTION 2: RELATED WORK
%% ============================================================
\section{Related Work}
\label{sec:related}

%\subsection{TOD and User Simulation}
\noindent\textbf{TOD and User Simulation.} 
Classical TOD systems adopt a modular NLU/DST/policy/NLG pipeline~\citep{chen2017survey};
%~\citep{young2013pomdp,chen2017survey}; 
MultiWOZ~\citep{budzianowski2018multiwoz} and its refinements~\citep{eric2019multiwoz21,zang2020multiwoz22,ye2022multiwoz24} remain the primary multi-domain benchmark, with a shift toward end-to-end architectures~\citep{qin2023etod_survey} pairing pretrained backbones~\citep{radford2019language,raffel2020t5} with LLMs in zero/few-shot settings~\citep{openai2023gpt4,hudecek2023are_llms_all_you_need}. User simulation alleviates data-collection cost: early works used rule- and agenda-based formulations with stack-like user goals~\citep{schatzmann2009hidden,li2016user_simulator}, while neural network based approaches improved diversity through seq2seq models~\citep{asri2016sequence}, neural NLG~\citep{kreyssig2018neural}, hierarchical latent-variable models~\citep{gur2018user}, multi-task models~\citep{kim2022multitask}, generative transformers with zero-shot ontology transfer~\citep{lin2022gentus}, and emotion-aware simulation~\citep{lin2023emous}.
Dialogue systems are widely deployed in e-commerce, with industrial systems such as AliMe~\citep{li2017alime} integrating intent recognition, knowledge retrieval, and multi-turn management at massive scale; financial dialogue systems impose additional compliance, accuracy, and professional-rigor requirements.

%\subsection{LLM-Based User Simulation}
\noindent\textbf{LLM-Based User Simulation.}
Prompt-based methods generate diverse user utterances without fine-tuning~\citep{davidson2023user_sim_llm,terragni2023icl_user_sim,almeida2024prompting_llm_usersim} but suffer from hallucinations, repetition, and incomplete goal fulfillment. Fine-tuning and verification-based approaches address these issues: DAUS fine-tunes on real dialogues for goal consistency~\citep{sekulic2024daus}; DuetSim uses a generator/verifier architecture to mitigate ``lost-in-the-middle'' failures~\citep{luo2024duetsim}; UGRO leverages LLM feedback to optimize smaller models~\citep{hu2023ugro}; utterance-level alignment sharpens response fidelity~\citep{luo2025utterance_alignment}. A parallel line incorporates profiles or personas~\citep{personallm_workshop}: USP infers \emph{implicit} user profiles and optimizes them via conditional SFT and RL with utterance- and conversation-level cycle consistency~\citep{wang2025usp}; mTAD composes trait-specific language models at decoding time for trait-conditioned simulation~\citep{ferreira2024mtad}; CSHI emphasizes controllable, human-involved simulation for conversational recommendation~\citep{zhu2024cshi}; \citet{zhang2025llm_usersim_rec} couple an LLM simulator with statistical preference models to train RL-based recommenders; and cognitive-architecture frameworks target general behavior simulation~\citep{wang2025user_behavior_sim,ren2024bases}---see \citet{balog2025user_sim_survey} for a comprehensive survey. \citet{frisch2025consistent_persona_rl} show that multi-turn RL reduces persona drift by over 55\%, directly motivating our GRPO design. These methods predominantly target general written-text scenarios; in contrast to USP's implicit profiles, \textsc{ProfileSim} conditions on explicit demographic--personality and business profiles reverse-engineered from in-domain call logs, leaving spoken-language modeling and business-process constraints---largely unaddressed in prior work---as its central focus.

%\subsection{Sales and Marketing Dialogue Systems}
Despite the commercial rise of voice AI in outbound sales, academic work on user simulation for outbound marketing remains sparse, focusing on the system side with little attention to user behavior under business-process constraints. We target user simulation specifically for financial outbound calling, combining business-intent and user-profile modeling with reinforcement learning under hierarchical dialogue- and utterance-level rewards to optimize spoken-language naturalness and long-range persona coherence; the resulting framework generates dialogues intended to support intent recognition training, agent evaluation, and cold-start deployment.

\begin{comment}
\subsection*{Summary}

Table~\ref{tab:comparison} provides a comparative overview of representative user simulation methods.

\begin{table}[htbp]
\centering
\caption{Comparison of representative user simulation methods.}
\label{tab:comparison}
\resizebox{\textwidth}{!}{%
\begin{tabular}{llll}
\toprule
\textbf{Method} & \textbf{Approach} & \textbf{Domain} & \textbf{Key Limitation} \\
\midrule
\citet{schatzmann2009hidden} & Agenda-based, statistical & General TOD & \citet{asri2016sequence} & Seq2Seq neural & General TOD & \citet{kreyssig2018neural} & Neural (NUS), NL generation & General TOD & Ontology-dependent \\
\citet{kim2022multitask} & Multi-task T5 & General TOD & No domain-specific modeling \\
\citet{lin2022gentus} & Generative transformer (GenTUS) & General TOD & No spoken-language features \\
\citet{lin2023emous} & Emotional user simulation & General TOD & \citet{davidson2023user_sim_llm} & LLM prompt engineering & General TOD & Hallucination, goal drift \\
\citet{sekulic2024daus} & LLM fine-tuning (DAUS) & General TOD & \citet{luo2024duetsim} & Dual-LLM (DuetSim) & General TOD & No colloquial/spoken features \\
\citet{frisch2025consistent_persona_rl} & Multi-turn RL for consistency & General (health, edu) & \citet{wang2025user_behavior_sim} & LLM agent simulation & Rec.\ systems, search & No dialogue-level optimization \\
\textbf{Ours} & \textbf{Profile-driven LLM + GRPO} & \textbf{Financial sales} & --- \\
\bottomrule
\end{tabular}%
}
\end{table}
\end{comment}

%% file: methodology.tex
%% ============================================================
%%  SECTION 3: METHODOLOGY
%% ============================================================
\begin{figure*}[t]
    \centering
    \includegraphics[width=\linewidth]{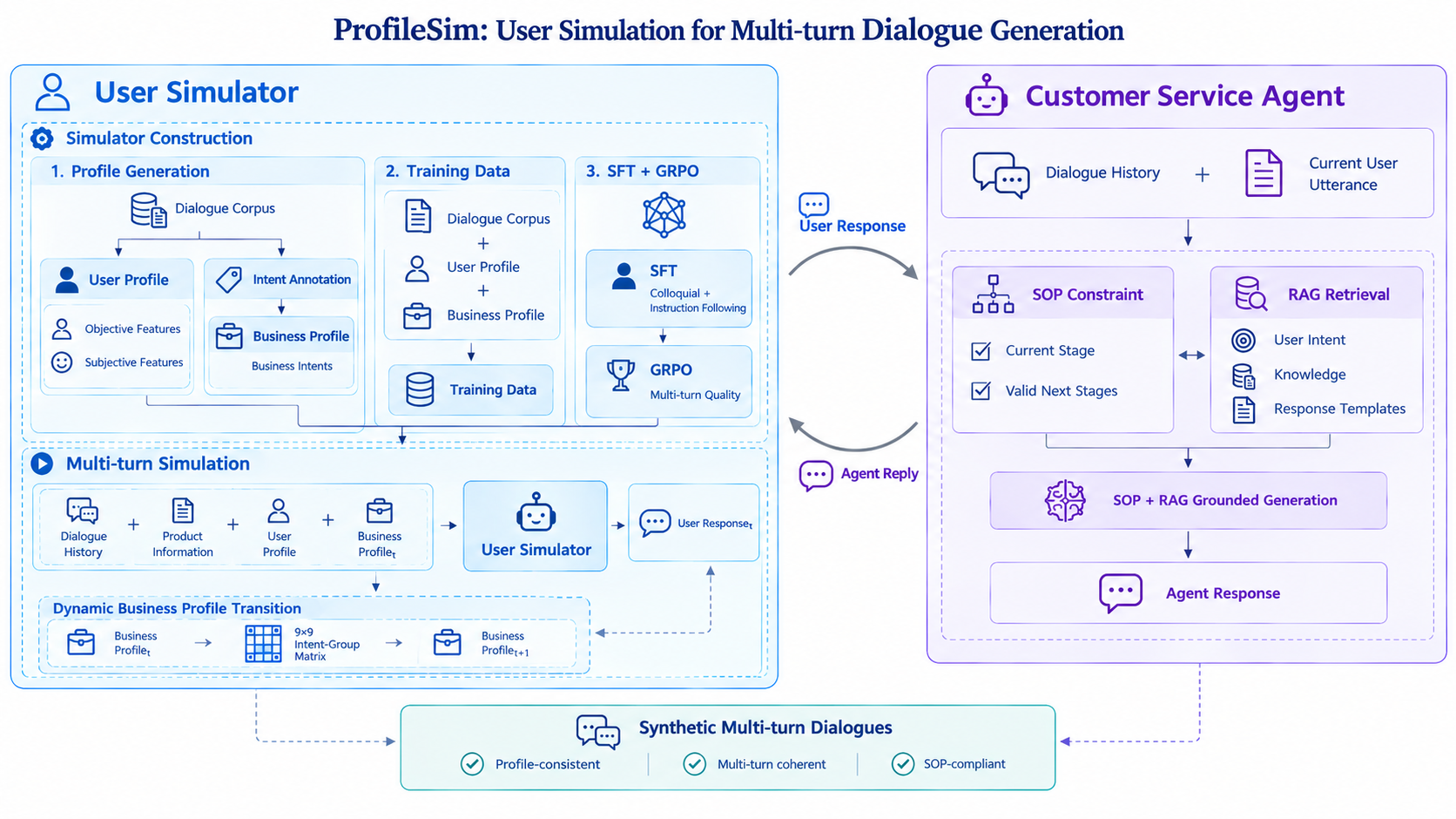}
    \caption{\textbf{\textsc{ProfileSim} couples a profile-conditioned user simulator with an SOP- and RAG-constrained agent to generate closed-loop dialogue data.} The User Simulator (left) is constructed through user/business profile extraction, supervised fine-tuning on real voice dialogue data, and GRPO with hierarchical dialogue- and utterance-level rewards. The Customer Service Agent (right) integrates SOP-constrained dialogue management with retrieval-augmented generation. The two agents interact to produce closed-loop multi-turn dialogue data.}
    \label{fig:framework}
\end{figure*}

\section{Methodology}
\label{sec:method}

This section presents the proposed \textsc{ProfileSim} framework in detail. As illustrated in Figure~\ref{fig:framework}, our system consists of two interacting agents: (1) a \emph{User Simulator} that generates realistic, profile-conditioned user responses, 
%for financial outbound calling scenarios, 
and (2) a \emph{Customer Service Agent} that interacts with the user simulator under structured business process constraints. Together, these components form a closed-loop simulation environment capable of generating high-quality multi-turn dialogue data for downstream training and evaluation.

At dialogue turn $t$, the User Simulator generates a user response $r_t$ conditioned on the dialogue history $h_t$, product information $x^{\text{prod}}$, user profile $p_{\text{user}}$, and business profile $p_{\text{biz},t}$. Simultaneously, the Customer Service Agent generates a service response $a_t$ based on the current dialogue stage $s_t$, retrieved knowledge $k_t$, and the dialogue context, and predicts the next stage $s_{t+1}$. 
%Formally:
%\begin{equation}
%    \textit{System} = \{\textit{UserSimulator},\; \textit{CustomerServiceAgent}\}
%\end{equation}

The User Simulator models the conditional generation:
\begin{equation}
    r_t = f_{\theta}(h_t, x^{\text{prod}}, p_{\text{user}}, p_{\text{biz},t}),
\end{equation}
where $f_{\theta}$ denotes the user simulation model parameterized by $\theta$. The Customer Service Agent models:
\begin{equation}
    (a_t, s_{t+1}) = g_{\psi}(h_t, x^{\text{prod}}, s_t, k_t),
\end{equation}
where $g_{\psi}$ is the agent model parameterized by $\psi$, $s_t$ is the current SOP stage, and $k_t$ is the knowledge retrieved at turn $t$.

The two agents interact in a turn-taking fashion to produce complete multi-turn dialogues. The resulting data is intended to support training intent recognition models, evaluating customer service systems, and bootstrapping deployment under cold-start conditions. The following subsections describe each component in detail.

\subsection{User Simulator Construction}
\label{sec:user_sim}

The construction of the User Simulator proceeds through three stages (Figure~\ref{fig:framework}). First, we clean raw voice dialogue data and extract structured user and business profiles. Second, these profiles and the cleaned corpus are used to construct SFT training data. Third, the base model undergoes SFT to acquire colloquial language generation and instruction-following capabilities, followed by Group Relative Policy Optimization (GRPO)~\citep{shao2024deepseekmath} to further improve utterance quality and multi-turn coherence.

\subsubsection{User Profile Extraction}
\label{sec:user_profile}

The user profile $p_{\text{user}}$ represents personal attributes and personality characteristics, serving as a control signal governing both the informational content and linguistic style of generated responses. We decompose user profiling into two complementary dimensions:

\paragraph{Objective features} encode observable demographic attributes, including gender, age range, occupation, income status, and language style (e.g., ``male, 40--55 years old, unstable income, direct and concise language style'').

\paragraph{Subjective features} are grounded in the Big Five personality model~\citep{john1999bigfive}, characterizing personality along five dimensions: Extraversion, Agreeableness, Conscientiousness, Emotional Stability, and Openness to Experience. This representation has been widely adopted in computational user modeling and dialogue personalization~\citep{mairesse2007personage,zheng2020personalized_dialogue}. For instance, a profile might specify ``low-to-medium Extraversion, medium Agreeableness, high Emotional Stability, low Openness,'' enabling the model to generate behaviorally consistent responses that reflect risk-averse tendencies in a loan consultation scenario.

\paragraph{Data preparation.}
We collected multi-turn dialogue data from real financial marketing outbound call scenarios. The original corpus comprises approximately 242,438 dialogue sessions from voice-based marketing calls. After desensitization, transcription error correction, and quality filtering, 72,296 valid dialogues were retained, averaging 7 turns per session.

Formally, given an original dialogue $D = \{(u_t, a_t)\}_{t=1}^{T}$, we correct transcription errors and obtain intent annotations $y_t \in \mathcal{Y}$ through manual labeling. The valid utterance set is $D_{\text{valid}} = \{u_t \mid y_t \in \mathcal{Y}_{\text{valid}}\}$.

\paragraph{Profile extraction model.}
Using valid utterances and intent labels, we employ an LLM to extract and summarize objective and subjective features, generating $p_{\text{user}}$ in natural language form. 
% The prompts used for objective feature extraction, subjective feature extraction, and natural-language profile generation are provided in Appendix~\ref{app:profile_prompts}. 
We constructed 10,000 ``dialogue--profile'' pairs and fine-tuned the Qwen3.5-9B base model~\citep{qwen35_2026} using LoRA~\citep{hu2022lora} to obtain a profile generation model $g_\phi$:
\begin{equation}
    \hat{p}_{\text{user}} = g_\phi(D)
\end{equation}
with the standard maximum likelihood training objective:
\begin{equation}
    \mathcal{L}_{\text{profile}}(\phi) = -\sum_{(D, p)} \log P_\phi(p \mid D)
\end{equation}

\subsubsection{Business Profile Schema}
\label{sec:biz_profile}

To enhance business controllability, we introduce a \emph{business profile} $p_{\text{biz}}$ that explicitly constrains the model's topical focus. Rather than constructing business constraints from prior domain knowledge, we adopt a \emph{reverse engineering} approach grounded in real dialogue data.

Starting from historical dialogues, we perform semantic analysis for each utterance and map natural language to structured business intent labels:
\begin{equation}
    u \rightarrow y_{\text{intent}}
\end{equation}
where $y_{\text{intent}}$ represents the business intent (e.g., ``Loan process consultation,'' ``Repayment method confirmation''). We then reverse-engineer these labels into natural language descriptions:
\begin{equation}
    y_{\text{intent}} \rightarrow p_{\text{biz}}
\end{equation}
where $p_{\text{biz}}$ is a natural language prompt describing the user's potential business concerns.

This pipeline of ``dialogue corpus $\rightarrow$ intent abstraction $\rightarrow$ profile generation'' achieves structured modeling of business knowledge from a data-driven perspective. Compared with using discrete labels directly, natural language profiles enhance the model's contextual understanding. In practice, the business profile is injected as an instruction prompt:
\begin{quote}
\textit{``Next, you may ask questions about $p_{\text{biz}}$. For example, you might inquire about the loan application process and pay close attention to the specific entry points and operation methods.''}
\end{quote}

This design explicitly guides the model toward specific domain concerns rather than unconstrained exploration, enhancing business relevance and controllability.

To further model the dynamic nature of user intentions in multi-turn conversations, we introduce a business profile transition mechanism that dynamically determines the business profile for the next turn.

\paragraph{Business Profile Transition.}

In our setting, the business profile space contains 73 fine-grained business intents, which are organized into 9 business intent groups according to their semantic and business relationships. Rather than manually specifying transition probabilities for all fine-grained intents, we estimate transition dynamics at the intent-group level from real customer-service dialogue trajectories.

Specifically, for two consecutive user turns, we record the current intent group $g_t$ and the subsequent intent group $g_{t+1}$, and estimate the conditional transition probability as
\begin{equation}
    P(g_{t+1}\mid g_t)
    =
    \frac{N(g_t,g_{t+1})}
    {\sum_{g'}N(g_t,g')}
\end{equation}
where $N(g_t,g_{t+1})$ denotes the number of observed consecutive intent-group pairs $(g_t,g_{t+1})$ in the real dialogue corpus. This yields a $9\times9$ intent-group transition matrix that captures empirical tendencies in how users shift between different business topics during conversations.

To account for users continuing to discuss the current topic across consecutive turns, we introduce a stay probability $p_{\mathrm{stay}}=0.1$. At each transition, the current intent group is retained with probability $p_{\mathrm{stay}}$, while the remaining probability mass is distributed among other intent groups in proportion to their empirical transition probabilities. This adjustment preserves the relative transition preferences between different groups while explicitly modeling the possibility of staying on the current topic.

The next business intent group is then sampled according to $g_{t+1}\sim\tilde{P}(\cdot\mid g_t)$. A fine-grained business profile $p_{\mathrm{biz},t+1}$ is subsequently sampled uniformly at random from the profiles within $g_{t+1}$. The resulting business profile is incorporated into the dialogue state for the next turn and provides explicit guidance for subsequent user utterance generation.

Overall, this transition mechanism preserves the empirical topic-transition patterns observed in real conversations while allowing the simulator to dynamically shift its business focus across turns.

\subsubsection{Supervised Fine-Tuning}
\label{sec:sft}

During supervised fine-tuning, we pursue two objectives: (1) adapting the model to colloquial expressions characteristic of outbound call scenarios~\citep{ji2024wavchat}, and (2) enabling conditional generation based on user and business profiles.

We formalize user simulation as a conditional text generation problem. Each training sample is a quintuple $(x^{\text{prod}}, h_t, p_{\text{user}}, p_{\text{biz}_t}, u_t)$, where the model generates the target response $u_t$ conditioned on product information $x^{\text{prod}}$, dialogue history $h_t$, and profiles. The learning objective is:
\begin{equation}
    P_\theta(u_t \mid x^{\text{prod}}, h_t, p_{\text{user}}, p_{\text{biz},t})
\end{equation}
with the autoregressive cross-entropy loss:
\begin{equation}
\begin{split}
    \mathcal{L}_{\text{SFT}}(\theta) = -\sum_{(x^{\text{prod}}, h_t, p_{\text{user}}, p_{\text{biz}}, u_t)} \log P_\theta(u_t \mid x^{\text{prod}}, \\ h_t, p_{\text{user}}, p_{\text{biz},t})
\end{split}
\end{equation}

\subsubsection{GRPO-Based Reinforcement Learning}
\label{sec:grpo}

After supervised fine-tuning, we further optimize the user simulator
with Group Relative Policy Optimization (GRPO)~\citep{shao2024deepseekmath}, which eliminates the need for a separate value network by estimating advantages through relative comparisons within groups of sampled responses. GRPO was originally proposed for enhancing mathematical reasoning~\citep{shao2024deepseekmath} and has since been adopted as the core RL algorithm in DeepSeek-R1~\citep{guo2025deepseekr1}.  The goal is to improve the simulator's ability to generate responses that are consistent with the target user profile while remaining coherent, natural, non-repetitive, and consistent with the evolving business context. During reinforcement learning, the user simulator is the trainable policy, whereas the customer-service agent is kept fixed and serves as the environment policy.

\paragraph{Multi-turn Dialogue Rollout.}
Given an initial dialogue context, we sample $G$ independent dialogues from the current user-simulator policy. At dialogue turn $t$, the context is defined as:
\begin{equation}
c_t =
\left(
x^{\mathrm{prod}},
h_t,
p_{\mathrm{user}},
p_{\mathrm{biz},t}
\right)
\label{eq:grpo_context}
\end{equation}
where $x^{\mathrm{prod}}$ denotes the product information, $h_t$ denotes the dialogue history, $p_{\mathrm{user}}$ denotes the target user profile, and $p_{\mathrm{biz},t}$ denotes the business profile associated with the current turn. Following the business profile transition mechanism described in Section~\ref{sec:biz_profile}, $p_{\mathrm{biz},t}$ is dynamically sampled according to the empirical intent-group transition distribution and the subsequent within-group profile sampling process. This enables the business focus to evolve across turns during the multi-turn rollout.

For the $i$-th dialogue, the user simulator generates
\begin{equation}
u_t^{(i)}
\sim
\pi_{\theta}
\left(
\cdot \mid c_t^{(i)}
\right),
\qquad
i=1,\ldots,G
\label{eq:grpo_policy}
\end{equation}
The generated user response is then provided to the fixed customer-service agent, which produces an agent response $a_t^{(i)}$. The dialogue history is subsequently updated with the new interaction and used to construct the context of the next turn. The rollout continues until a terminal condition is reached or the maximum number of turns is exceeded.

Accordingly, each group consists of $G$ independently sampled
multi-turn dialogues:
\begin{equation}
\mathcal{G}
=
\left\{
\mathcal{T}^{(1)}
\ldots,
\mathcal{T}^{(G)}
\right\}
\label{eq:grpo_group}
\end{equation}
\begin{equation}
\mathcal{T}^{(i)}
=
\left\{
(u_1^{(i)},a_1^{(i)}),
\ldots,
(u_{T_i}^{(i)},a_{T_i}^{(i)})
\right\}
\label{eq:dialogue}
\end{equation}
where $T_i$ is the number of turns in each dialogue.

Full-dialogue rollouts are required in our setting because the reward function includes a profile-consistency component computed from the cumulative user responses. Thus, the quality of a generated response cannot be fully evaluated independently of the subsequent dialogue trajectory.

\paragraph{Hierarchical Reward Design.}
We design a hierarchical reward function to evaluate the generated dialogues at both the dialogue and utterance levels. The dialogue-level reward captures long-term consistency of the simulated user, whereas the utterance-level reward evaluates the quality of individual user responses.

For dialogue $\mathcal{T}^{(i)}$, the dialogue-level reward is defined by the consistency between the target user profile and the profile implied by the complete sequence of generated user responses:
\begin{equation}
r_{\mathrm{profile}}^{(i)}
=
\operatorname{sim}
\left(
\hat{p}_{\mathrm{user}}^{(i)},
p_{\mathrm{user}}^{(i)}
\right)
\label{eq:profile_reward}
\end{equation}
where $\hat{p}_{\mathrm{user}}^{(i)}$ is extracted from the cumulative user responses in dialogue $i$, and $\operatorname{sim}(\cdot,\cdot)$ measures their cosine similarity. The resulting profile-consistency score lies in the range $[0,1]$.

At each turn, we additionally evaluate the quality of the individual user response. The utterance-level reward is defined as:
\begin{equation}
\begin{aligned}
r_{\mathrm{utterance},t}^{(i)}
={}&
\alpha_1 r_{\mathrm{coh},t}^{(i)}
+\alpha_2 r_{\mathrm{nonrep},t}^{(i)} \\
&+\alpha_3 r_{\mathrm{natural},t}^{(i)}
+\alpha_4 r_{\mathrm{biz},t}^{(i)}
\end{aligned}
\label{eq:utterance_reward}
\end{equation}
where $r_{\mathrm{coh}}$ measures contextual coherence, $r_{\mathrm{nonrep}}$ penalizes repetitive responses, $r_{\mathrm{natural}}$ measures linguistic naturalness and colloquialism, and $r_{\mathrm{biz}}$ measures consistency with the business profile.

We set $\alpha_1=0.30, \alpha_2=0.15, \alpha_3=0.25, \alpha_4=0.30.$ This weighting places greater emphasis on contextual coherence, business consistency, and linguistic naturalness, while assigning a smaller but complementary weight to non-repetition.

Unless otherwise noted, Gemini~3.1 Pro serves as the LLM-as-judge~\citep{zheng2024judging}.
% , with rubric prompts provided in Appendix~\ref{app:reward_prompts}. 
For coherence and naturalness, the evaluator assigns a score on a 1--5 scale, which is normalized to $[0,1]$ as:
\begin{equation}
\operatorname{norm}(s)=\frac{s-1}{4}
\label{eq:reward_normalization}
\end{equation}
The non-repetition and business-consistency rewards are binary and therefore already lie in $[0,1]$. For non-repetition, the cosine similarity between the current user response and the preceding user response is computed; if the similarity exceeds $0.8$, the reward is set to $0$, and otherwise it is set to $1$.

\paragraph{Turn-Level Reward.}
The dialogue-level and utterance-level objectives are combined into the reward used for policy optimization. Specifically, for each turn we define:
\begin{equation}
r_t^{(i)}
=
\lambda_d r_{\mathrm{profile}}^{(i)}
+
\lambda_u r_{\mathrm{utterance},t}^{(i)}
\label{eq:total_turn_reward}
\end{equation}
where $\lambda_d=0.5, \lambda_u=0.5.$

Because the profile reward is computed from the cumulative user responses, the same dialogue-level profile-consistency score is associated with all turns of a dialogue, while the utterance-level reward can vary across turns. Consequently, the resulting reward sequence for dialogue $i$ is
\begin{equation}
\mathbf{r}^{(i)}
=
\left(
r_1^{(i)},
\ldots,
r_{T_i}^{(i)}
\right)
\label{eq:reward_sequence}
\end{equation}

If the reward model instead produces a scalar dialogue-level reward, the implementation broadcasts this scalar to all turns:
\begin{equation}
r_t^{(i)}=r^{(i)},
\qquad
t=1,\ldots,T_i
\label{eq:scalar_reward_broadcast}
\end{equation}
Therefore, the same GRPO procedure supports both scalar dialogue-level rewards and utterance-level reward sequences.

\paragraph{Group-Relative Advantage Estimation.}
To obtain a group-relative learning signal, each dialogue is first represented by the mean of its turn-level rewards:
\begin{equation}
R_i
=
\frac{1}{T_i}
\sum_{t=1}^{T_i}
r_t^{(i)}
\label{eq:dialogue_reward}
\end{equation}
For the $G$ dialogues sampled from the same initial context, we compute the group mean and standard deviation as
\begin{equation}
\mu_{\mathcal{G}}
=
\frac{1}{G}
\sum_{i=1}^{G}R_i
\label{eq:group_mean}
\end{equation}

\begin{equation}
\sigma_{\mathcal{G}}
=
\sqrt{
\frac{1}{G}
\sum_{i=1}^{G}
\left(
R_i-\mu_{\mathcal{G}}
\right)^2
}
\label{eq:group_std}
\end{equation}
For numerical stability, $\sigma_{\mathcal{G}}$ is lower-bounded by a small constant $\epsilon$.

Rather than assigning the same dialogue-level advantage to every turn, we preserve the variation of rewards across turns. Specifically, the advantage of turn $t$ in dialogue $i$ is
\begin{equation}
A_{i,t}
=
\frac{
r_t^{(i)}-\mu_{\mathcal{G}}
}{
\sigma_{\mathcal{G}}+\epsilon
}
\label{eq:turn_level_advantage}
\end{equation}

This formulation provides fine-grained turn-level credit assignment while retaining the group-relative normalization of GRPO. In particular, turns whose rewards are higher than the group mean receive positive advantages, whereas turns with lower rewards receive negative advantages. Importantly, the group statistics are computed from the dialogue-level mean rewards $R_i$, rather than directly from all individual turn rewards.

When only a scalar dialogue-level reward is available, the reward is broadcast to every turn, and Eq.~\ref{eq:turn_level_advantage} becomes
\begin{equation}
A_{i,t}
=
\frac{
R_i-\mu_{\mathcal{G}}
}{
\sigma_{\mathcal{G}}+\epsilon
}
\label{eq:scalar_grpo_advantage}
\end{equation}
which recovers the standard dialogue-level group-relative signal.

Moreover, the average of the turn-level advantages within each dialogue is equal to its corresponding dialogue-level group-relative advantage:
\begin{equation}
\frac{1}{T_i}
\sum_{t=1}^{T_i}A_{i,t}
=
\frac{
R_i-\mu_{\mathcal{G}}
}{
\sigma_{\mathcal{G}}+\epsilon
}
\label{eq:advantage_consistency}
\end{equation}
Thus, the proposed turn-level formulation preserves the original dialogue-level GRPO signal while allowing different turns within the same dialogue to receive different learning signals.

\subsection{Customer Service Agent Construction}
\label{sec:cs_agent}

To evaluate the user simulator in realistic settings and complete the simulation loop, we construct a customer service agent providing a structured interactive environment.

\subsubsection{SOP-Constrained Dialogue}

In real financial sales scenarios, conversations are strictly constrained by Standard Operating Procedures (SOPs)~\citep{kulkarni2025sop_agent}. We model the customer service dialogue as a decision process governed by procedural constraints rather than free-form generation.

We represent the SOP as a directed graph $\mathcal{G} = (\mathcal{V}, \mathcal{E})$, where each node $v \in \mathcal{V}$ corresponds to a dialogue stage (e.g., opening remarks, product introduction, channel guidance) and each edge $e \in \mathcal{E}$ represents an allowed transition. At each turn, the agent may only remain at the current stage or transition to a legal subsequent stage, confining the dialogue within a structured procedural space. This graph-based representation ensures adherence to the prescribed business workflow regardless of user input diversity.

\subsubsection{Retrieval-Augmented Generation}

While the SOP graph provides procedural structure, accurate product information is needed for professional responses. We employ Retrieval-Augmented Generation (RAG)~\citep{lewis2020rag,gao2024rag_survey} to ground the agent's responses in domain knowledge.

At each turn, the system retrieves three types of information from the knowledge base: (1) the user's business intent recognition result, (2) candidate next dialogue stages, and (3) dialogue templates for each candidate stage. These elements are filtered using the current SOP state, retaining only procedurally valid candidates. This retrieval-and-filter mechanism ensures both flexibility in responding to diverse queries and consistency in adhering to business processes.

\subsubsection{Structured Input and Output}

The agent receives as input: historical dialogue, product information, current stage, SOP-filtered candidate stages, and retrieved knowledge. These inputs are organized into a unified prompt, optionally augmented with chain-of-thought reasoning~\citep{wei2022cot} for explicit process reasoning.

The output is structured JSON containing two fields: the predicted next stage and the corresponding response. This format facilitates integration with downstream modules such as dialogue state tracking and execution engines, enhancing engineering usability and deployability.

%% file: experiments.tex
%% ============================================================
%%  SECTION 4: EXPERIMENTS
%% ============================================================
\section{Experiments}
\label{sec:experiments}

In this section, we present a comprehensive experimental evaluation of the proposed user simulation framework. We first describe the experimental settings, including training details, the customer service simulation environment, and baseline models (Section~\ref{sec:settings}). We then analyze distributional alignment with real-world conversations using perplexity-based evaluation across different model configurations (Section~\ref{sec:distribution_alignment}). Next, we report the generated dialogue dataset statistics (Section~\ref{sec:dataset}), and conduct a multi-dimensional evaluation of simulator fidelity (Section~\ref{sec:simulator_fidelity}).

\subsection{Experimental Settings}
\label{sec:settings}

\paragraph{Training details.}
Our full set of training data consists of 72,296 multi-turn dialogues collected from real outbound call scenarios, comprising 432,457 utterances across multiple financial domains, including insurance and wealth management.

We first trained a user profile extraction model to obtain structured user profiles. We sampled 10,000 multi-turn dialogues from the training corpus and constructed profile annotations using GPT-5~\citep{openai2025gpt5}. To assess the quality of the generated profiles, we randomly sampled 1,000 dialogue--profile pairs for manual validation. The annotators evaluated whether each profile attribute was supported by the corresponding dialogue and whether the resulting profile was coherent and complete. 95.4\% of the evaluated profiles were judged valid, suggesting that the automatically constructed profiles provide high-quality supervision for subsequent user-simulation training. 

Based on the Qwen3.5-9B base model, we fine-tuned the profile extraction model with LoRA for 3 epochs on 2 NVIDIA A100 GPUs, requiring approximately 4 hours. The trained profile extraction model was then applied to the remaining dialogues. The extracted user profiles were subsequently used to construct the training data for the user simulation model. 

We used SFT to train the user simulation model based on the Qwen3.5-9B base model by sampling 97,126 utterances from the training dataset. We employed LoRA fine-tuning on 2 NVIDIA A100 GPUs for 3 epochs, with a maximum output sequence length of 1,024 tokens and a learning rate of $2.0 \times 10^{-4}$. The SFT training required approximately 14 hours.

For the GRPO-based preference optimization stage (Section~\ref{sec:grpo}), we performed LoRA fine-tuning with a rollout group size of $G = 8$. We randomly sampled 4,000 user profiles from the 72,296 profiles extracted from the real outbound call dialogues. Since GRPO was conducted in an online reinforcement learning setting, only the user profiles were used as the initial conditions for interaction. During training, the user simulation model interacted with a separate customer service model to dynamically generate multi-turn dialogue trajectories, which were used for subsequent policy optimization. Training was conducted on 8 NVIDIA H200 GPUs for 2 epochs and required approximately 20 hours.

\paragraph{Customer service simulation.}
To construct the interactive environment for the user simulator, we employed DeepSeek-R1~\citep{guo2025deepseekr1} as the customer service agent backbone. The agent integrates Retrieval-Augmented Generation (RAG)~\citep{lewis2020rag} with the SOP-constrained dialogue management framework described in Section~\ref{sec:cs_agent} to generate contextually grounded and procedurally compliant customer service responses.

\paragraph{Baselines.}
We compare the proposed user simulation model against multiple large language models: DeepSeek-R1~\citep{guo2025deepseekr1}, DeepSeek-V3~\citep{deepseek2024v3}, GPT-5, and Qwen3-Max~\citep{qwen3max_2025}. All baselines are evaluated in a zero-shot setting without task-specific fine-tuning. Importantly, the baselines and our model are given exactly the same prompt template and input information, including the user profile, business profile, product information, and dialogue history. Thus, the only difference is the underlying model and whether it has undergone our task-specific training, providing a controlled comparison of the benefit of the proposed training approach.

We also considered direct comparison with existing purpose-built user models, such as UserLM~\citep{naous2026flipping}. However, its training and evaluation setting differs substantially from ours. UserLM is designed to simulate general human users in interactions with AI assistants and is evaluated primarily in open-domain multi-turn scenarios, including coding and mathematical conversations. In contrast, our model targets Chinese outbound business conversations, where the simulator must generate realistic colloquial spoken responses while being explicitly conditioned on domain-specific user profiles, business profiles, product information, and dialogue context. As a result, directly applying UserLM would introduce substantial mismatches in language, dialogue domain, and conditioning setting, making a direct comparison less informative. We therefore use strong general-purpose LLMs under an identical prompting setup as our primary baselines, allowing us to isolate and evaluate the benefit of task-specific training for user simulation in the target scenario.

\subsection{Distributional Alignment Analysis}
\label{sec:distribution_alignment}

To evaluate how well different model configurations align with the distribution of real user conversations, we conduct distributional alignment analysis using perplexity (PPL), a standard intrinsic metric for measuring the uncertainty of language models over human-written text.

Given a test set containing a total of \(N\) tokens, the perplexity of a model \(P_\theta\) is computed as:

\begin{equation}
\mathrm{PPL}(P_\theta)
=
\exp\left(
-\frac{1}{N}
\sum_{i=1}^{N}
\log P_\theta(x_i \mid x_{<i})
\right)
\end{equation}

Lower perplexity indicates that the model assigns higher probability to real user responses, suggesting better alignment with the underlying distribution of human conversations.

To analyze the contribution of different components in the proposed framework, we compare the following model configurations:

\begin{itemize}
    \item \textbf{Qwen3.5-9B\_SFT}: Qwen3.5-9B fine-tuned with SFT on the training data, while the prompt excludes both the business profile and user profile; only dialogue history is used for conditioning.
    
    \item \textbf{+ Business Profile}: Qwen3.5-9B fine-tuned via SFT on the training data with only business profile conditioning.

    \item \textbf{+ User Profile}: Qwen3.5-9B fine-tuned via SFT on the training data with only user profile conditioning.

    \item \textbf{Full SFT}: Qwen3.5-9B fine-tuned on the training data with both user and business profile conditioning.
    
    \item \textbf{Full Model}: The complete proposed framework, which further applies GRPO-based preference optimization on top of the Full SFT model using the hierarchical reward function.
\end{itemize}

All configurations are evaluated on the same held-out test set collected from real-world marketing conversations. For each configuration, perplexity is computed conditioned on the corresponding profile inputs and dialogue history. All models share identical decoding settings and evaluation protocols to ensure fair comparison.

Table~\ref{tab:ppl_ablation} reports the perplexity results across different model configurations.

\begin{table}[h]
\centering
\small
\caption{\textbf{Profile conditioning drives distributional alignment, and GRPO preserves it.} Distributional alignment measured by perplexity (PPL). Lower values indicate better alignment with real user conversation distributions. The best result is highlighted in \textbf{bold}.}
\label{tab:ppl_ablation}
\begin{tabular}{lc}
\toprule
\textbf{Model} & \textbf{PPL} $\downarrow$ \\
\midrule 
Qwen3.5-9B\_SFT & 58.48 \\
+ Business Profile & 43.04 \\
+ User Profile & 22.24 \\
Full SFT & 17.52 \\
\textbf{Full Model} & \textbf{17.49} \\
\bottomrule
\end{tabular}
\end{table}

\paragraph{Effect of User and Business Profiles.}
We first compare all profile-conditioned variants against \textbf{Qwen3.5-9B\_SFT} (PPL = 58.48), which is domain-adapted via SFT on real outbound-call dialogues but receives only dialogue history at inference time, without business or user profile prompts. This isolates the gain attributable to \emph{explicit profile conditioning} beyond task-specific fine-tuning alone. Introducing \textbf{business profile} conditioning (+ Business Profile) reduces PPL to 43.04 ($\downarrow$15.44), showing that topical business guidance helps the model concentrate on scenario-relevant concerns and marketing interaction patterns rather than generic user language. \textbf{User profile} conditioning yields a substantially larger improvement (+ User Profile: 22.24, $\downarrow$36.24), indicating that persona-specific preferences and behavioral cues are the dominant factor for matching the distribution of real user responses in this setting. Notably, user profile alone outperforms business profile alone by a wide margin (22.24 vs.\ 43.04), suggesting that personalization signals are more informative than business-scope constraints when modeling \emph{who} is speaking. Combining both profiles in \textbf{Full SFT} further lowers PPL to 17.52, demonstrating that business and user profiles are \textbf{complementary}: the user profile captures individualized behavior, while the business profile refines domain-appropriate intent expression on top of that persona grounding. Together, these results show that SFT alone is insufficient for high-fidelity user simulation in marketing-oriented dialogues; structured profile conditioning---especially user profile---is critical for distributional alignment with real outbound user utterances.

\paragraph{Effect of GRPO Preference Optimization.}
Building upon the \textbf{Full SFT} configuration, the \textbf{Full Model} further reduces PPL from 17.52 to 17.49 after applying GRPO-based preference optimization with the hierarchical reward function. Although the absolute improvement is small, this result shows that GRPO does not degrade the model's distributional alignment with real user conversations, while providing additional optimization beyond supervised fine-tuning. More importantly, the limited PPL gain suggests that the major improvement in token-level distributional alignment is already obtained through SFT with explicit business and user profile conditioning. This is expected because PPL primarily measures token-level likelihood and is therefore insensitive to several behavioral properties that are central to user simulation, such as long-range dialogue coherence, persona consistency, realistic response strategies, and the ability to generate challenging interactions for downstream agents.

\subsection{Generated Dialogue Dataset}
\label{sec:dataset}

We generated a multi-turn dialogue dataset through online interaction between the user simulator and the customer service agent in financial product recommendation and sales scenarios. The customer service agent initiates each conversation using a predefined opening template, while the user simulator dynamically generates responses conditioned on the user profile and business profile. Across turns, the business profile evolves through the transition mechanism described in Section~\ref{sec:biz_profile}, with empirical transition probabilities estimated from real outbound call data, allowing the simulator to shift between different business intents while maintaining consistency with the current dialogue state.

The resulting dataset contains \textbf{3,000 high-quality dialogue sessions} and approximately \textbf{20,000 user--agent utterance pairs}, with an average of \textbf{7 turns per dialogue}. Each dialogue terminates when either the maximum number of interaction turns is reached or the user expresses an intent to end the conversation.The generated dialogues cover both scenario-general and scenario-specific business intents, with \textbf{58\% of user utterances} corresponding to scenario-specific intents. This indicates that the generated data contains a high proportion of business-relevant content and is consistent with the characteristics of real outbound call scenarios. A representative dialogue example is provided in Appendix~\ref{app:dialogue_example}.

\subsection{Simulator Fidelity Evaluation}
\label{sec:simulator_fidelity}

We evaluate simulator fidelity from three complementary perspectives:
(1) \emph{behavioral realism},
(2) \emph{controllability and role consistency},
and (3) \emph{downstream utility}.
Behavioral realism measures whether generated users resemble real outbound callers in conversational style and interaction patterns.
Controllability and role consistency evaluate whether the simulator remains faithful to conditioned user personas and business guidance.
Finally, downstream utility evaluates whether simulated users effectively expose weaknesses in a fixed customer service agent under realistic long-tail business scenarios.

Unless otherwise noted, Gemini~3.1 Pro serves as the LLM-as-judge. For metrics whose evaluation rubrics are also used as reward functions during GRPO training, we directly reuse the corresponding rubrics defined in Appendix~\ref{app:rubric_naturalness} and Appendix~\ref{app:rubric_business}, respectively. The remaining evaluation rubrics are provided in Appendix~\ref{app:rubrics}.
Human spot-checks on 500 stratified samples per metric show substantial agreement with the automated judge, with Cohen's $\kappa \geq 0.72$.

\subsubsection{Behavioral realism}

\paragraph{Colloquial naturalness.}
Colloquial naturalness measures whether generated user utterances resemble spontaneous spoken telephony conversations rather than written, scripted, or templated text. We directly reuse the same LLM-based evaluation rubric as the $r_{\mathrm{natural}}$ reward used during GRPO training (Appendix~\ref{app:rubric_naturalness}). We report the mean score across all evaluated user utterances, with higher scores indicating more natural and human-like conversational expression.

\paragraph{Utterance length.}
Outbound users typically produce short, reactive replies. We report the \textbf{mean character count} (\#chars) of user utterances per corpus; \textbf{lower} values are preferred for telephony realism, subject to retaining intelligible content.

\subsubsection{Controllability and role consistency}

\paragraph{User profile adherence.}
User profile adherence measures whether the simulated user behavior remains consistent with the predefined persona throughout the entire multi-turn interaction. The target persona is specified by structured attributes such as risk preference, prior product experience, communication style, and cooperation willingness. Concretely, we apply the user profile extraction model to the user's \textbf{cumulative responses} in each session to recover the profile implied by the generated behavior. We then compute the \textbf{cosine similarity} between the extracted profile representation and the target profile, yielding a score in $[0, 1]$. We report the mean score averaged over all dialogue sessions; \textbf{higher} values indicate stronger persona consistency throughout the interaction.

\begin{table*}[t]
\centering
\small
\caption{\textbf{Our full model attains the highest colloquial naturalness and user profile adherence and the lowest agent score among all evaluated user generators.} Comparative evaluation of simulated dialogue corpora (3,000 sessions per user generator; identical customer service agent and sampling protocol). Colloq.\ is a 1--5 LLM-judge rating ($\uparrow$ better). Length is mean user utterance length in characters ($\downarrow$ preferred). User Profile is the cosine similarity between the extracted and target user profiles, expressed as a percentage ($\times 100$) for readability ($\uparrow$ better). Business Profile is the intent-compliance rate in \% ($\uparrow$ better). Agent score is the mean judge rating on long-tail intent turns ($\downarrow$ indicates higher stress on the fixed agent). \textbf{Bold} marks our full model.}
\label{tab:dataset_comparison}
\begin{tabular}{lcccccccc}
\toprule

& \multicolumn{2}{c}{\textbf{Behavioral Realism}}
& \multicolumn{2}{c}{\textbf{Controllability \& Consistency}}
& \textbf{Utility} \\

\cmidrule(lr){2-3}
\cmidrule(lr){4-5}
\cmidrule(lr){6-6}

\textbf{User Generator}
& Colloq.
& Length
& User Profile
& Business Profile
& Agent Score \\

& $\uparrow$
& $\downarrow$
& $\uparrow$
& $\uparrow$
& $\downarrow$ \\

\midrule

DeepSeek-R1
& 3.30 & 120.89
& 84.05 & 32.46%
& 4.24 \\

DeepSeek-V3
& 3.22 & 153.96
& 83.42 & 16.67%
& 4.18 \\

GPT-5
& 4.21 & 44.54
& 83.18 & 62.01%
& 4.27 \\

Qwen3-Max
& 4.03 & 90.82
& 83.20 & 52.38%
& 4.45 \\

Ours Full SFT
& 4.46 & 11.62
& 84.29 & 52.21%
& 3.56 \\

\textbf{Ours Full Model}
& \textbf{4.97} & \textbf{11.14}
& \textbf{92.55} & \textbf{52.25}
& \textbf{3.49} \\

\bottomrule
\end{tabular}
\end{table*}

\paragraph{Business profile compliance.}
Recall that the business profile $p_{\text{biz}}$ is a natural-language instruction derived from real dialogues via intent abstraction ($u \rightarrow y_{\text{intent}} \rightarrow p_{\text{biz}}$), describing the user's plausible business concerns and the topical scope that the simulator should follow. It is injected at generation time to improve \emph{business controllability} by constraining the user's topical focus. To evaluate whether generated utterances adhere to the conditioned business profile, we directly reuse the same LLM-based evaluation rubric as the $r_{\mathrm{biz}}$ reward used during GRPO training (Appendix~\ref{app:rubric_business}). We report the mean score across all user utterances; higher values indicate stronger adherence to the conditioned business guidance.

\subsubsection{Downstream utility}

\paragraph{Downstream agent stress.}
Beyond intrinsic measures of user-simulation fidelity, we evaluate the downstream utility of generated user behaviors by measuring how well they expose weaknesses of a customer service agent. We focus on \textbf{long-tail business intents}, defined as intents occupying the bottom 20\% of frequency in the real training corpus (e.g., niche fee rules or rare objection types). Thus, the evaluated cases are grounded in the empirical distribution of real user conversations rather than artificially constructed adversarial inputs. Holding the agent model and RAG index fixed, we measure the agent's response quality on generated user turns corresponding to these long-tail intents using the same LLM-as-judge protocol. Lower agent scores indicate that the generated behaviors are more effective at exposing weaknesses of the fixed agent, providing greater downstream training value. 

\subsubsection{Results and analyses}

Table~\ref{tab:dataset_comparison} presents a comprehensive comparison between our user simulator and several strong general-purpose LLMs under the same customer service agent and sampling protocol. Overall, \emph{Ours Full Model} achieves the strongest performance across most behavioral realism and controllability metrics. In particular, it obtains the highest colloquial score of 4.97, substantially outperforming both general-purpose LLMs and the SFT-only variant (4.46). Meanwhile, its average utterance length is only 11.14 characters, indicating that the generated responses are concise and more consistent with realistic customer-side interactions. These results suggest that our simulator is less prone to the verbose and written-style responses commonly observed in general-purpose LLMs.

The most notable improvement is observed in profile grounding. \emph{Ours Full Model} achieves a profile score of 0.9255, compared with 0.8429 for \emph{Ours Full SFT} and at most 0.8405 among the general-purpose LLMs. This substantial improvement indicates that the full training pipeline enables the simulator to better preserve and express user-specific characteristics throughout the dialogue, which is important for generating behaviorally consistent simulated customers. In contrast, the general-purpose LLMs exhibit relatively similar profile scores around 0.83--0.84, suggesting that prompting alone is insufficient to reliably maintain the desired user profile.

We further evaluate the generated dialogues from the perspective of downstream stress testing. A lower \emph{Agent Score} indicates that the generated user interactions impose greater difficulty on the fixed customer service agent. \emph{Ours Full Model} achieves the lowest score of 3.49, slightly improving over \emph{Ours Full SFT} (3.56) and substantially outperforming all general-purpose LLMs (4.18--4.45). This result is particularly important: although general-purpose LLMs can generate fluent responses, their interactions are relatively easy for the customer service agent to handle. In contrast, our simulator produces more challenging long-tail intent interactions, making the generated corpus more suitable for downstream robustness and stress testing.

For business profile compliance, \emph{Ours Full Model} achieves 52.25\%, which is comparable to \emph{Ours Full SFT} (52.21\%) and Qwen3-Max (52.38\%), but lower than GPT-5 (62.01\%). This relatively modest compliance rate should be interpreted together with the colloquial naturalness of the generated utterances. General-purpose LLMs tend to express the conditioned business profile more explicitly and completely, which makes their outputs more likely to satisfy a semantic compliance criterion. However, such explicit formulations can be less consistent with the concise, elliptical, and spontaneous expressions commonly observed in real user conversations. As illustrated in Appendix~\ref{app:business_compliance_cases}, our model may express the same business intent through shorter and more implicit utterances, whereas general-purpose LLMs tend to explicitly restate the target business intent and its surrounding context.

Therefore, business profile compliance alone does not fully capture the quality of user simulation. A high compliance score can partly result from explicitly verbalizing the conditioned business intent, while realistic user behavior often involves implicit, abbreviated, or context-dependent expressions. Taken together, these results suggest that the gains of \emph{Ours Full Model} are not primarily reflected in explicit business profile compliance, but rather in its ability to jointly model naturalistic user behavior, controllable user characteristics, and challenging multi-turn interactions.

%% file: conclusion.tex
%% ============================================================
%%  SECTION 5: CONCLUSION
%% ============================================================
\section{Conclusion}
\label{sec:conclusion}

We presented \textsc{ProfileSim}, a profile-driven user simulation framework for financial marketing outbound calls that jointly models user profiles and business profiles to enable fine-grained control over simulated behavior. The framework combines supervised fine-tuning on real voice dialogue data with GRPO-based preference optimization under hierarchical rewards at the dialogue and utterance levels, and is paired with an SOP- and RAG-constrained customer service agent to produce closed-loop multi-turn dialogues. On financial marketing data, \textsc{ProfileSim} achieves the highest colloquial naturalness and user profile adherence among all evaluated user generators, while producing substantially shorter and more concise user utterances. It also imposes greater stress on the fixed customer service agent, as reflected by the lowest agent score on long-tail intents, indicating that the generated dialogues are more challenging and behaviorally realistic. Ablation results further show that explicit user and business profile conditioning substantially improves distributional alignment, while GRPO provides additional gains in behavioral fidelity beyond SFT. The generated dialogue data is designed to support intent recognition training, agent evaluation, and rapid cold-start deployment.

%% file: limitations.tex
\section*{Limitations}

The user profile representation is primarily static and does not capture dynamic shifts in attitude or emotion during a conversation. In addition, the framework is currently validated on a single domain (financial marketing outbound calls), and cross-domain experiments would be needed to further assess its generality. Promising directions include speech-level paralinguistic modeling~\citep{shim2026spokenus}, online co-evolutionary training of the simulator and agent, and cross-domain transfer.

%% file: citations.tex
%% ============================================================
%%  BIBLIOGRAPHY
%%  Note: for EMNLP/ACL submissions the canonical bibliography
%%  source is BibTeX with style acl_natbib. The thebibliography
%%  block below uses author-year labels so that \citet / \citep
%%  render correctly via natbib. For camera-ready, consider
%%  migrating entries to custom.bib and using:
%%    \bibliography{custom}
%%    \bibliographystyle{acl_natbib}
%% ============================================================
\bibliographystyle{acl_natbib}

%% file: appendix.tex
%% ============================================================
%%  APPENDIX
%%  Additional examples and supplementary materials.
%%  This file is included from the main manuscript via:
%%    \appendix
%%    \input{appendix}
%% ============================================================
\begin{CJK}{UTF8}{gbsn}

\section{User Profile Construction Prompts}
\label{app:profile_prompts}

\subsection{Objective Feature Extraction}
\label{app:objective_feature}

\begin{center}
\begin{promptbox}{Prompt for Objective Feature Extraction}
\label{prompt:objective_feature}
\small

\textbf{角色：}
专业的用户画像分析专家，专门根据用户在对话中的实际表达提取和总结用户的客观特征。

\medskip

\textbf{任务：}
根据给定的用户话语，提取与用户画像相关的客观特征，并按照指定字段生成结构化的用户画像。提取的特征应以用户话语中明确或合理支持的信息为依据。

\medskip

\textbf{字段描述：}
\begin{itemize}[leftmargin=*, nosep]
\item \textbf{性别：}此类别定义角色的性别认同。它可以包括男性、女性、非二元性别或任何其他性别认同。

\item \textbf{年龄：}此类别包含角色的具体年龄或估计年龄范围。

\item \textbf{地点：}这包括角色居住或工作的地理区域。它可以像国家或大洲一样宽泛，也可以像城市或社区一样具体。

\item \textbf{职业：}这详细描述了角色当前工作和工作经验。它包括行业、职位和工作年限，从而深入了解角色的技能、日常活动和职业挑战。

\item \textbf{教育：}这涵盖了角色的教育背景，包括最高学历、学习领域和重要的学校经历。

\item \textbf{家庭关系：}此类别概述了角色与家庭成员的关系，包括父母、兄弟姐妹、子女和其他亲属。

\item \textbf{日常习惯：}指人物角色定期进行的行为或活动。这些可以包括晨间例行活动、锻炼计划、固定的饮食习惯或定期的社交活动。

\item \textbf{社交关系：}这涉及人物角色与直系亲属以外的人员和团体的互动，包括朋友、同事或社区团体。此类别可以帮助我们了解人物角色的社交网络、支持系统和冲突处理策略。

\item \textbf{人格特质：}这包括构成个人性格的内在属性，例如内向或外向、乐观或悲观、随性或计划性。

\item \textbf{其他经历：}这是一个涵盖所有情况的类别，用于描述其他无法完全归入上述类别的重要经历。

\item \textbf{目标或计划：}此类别概述人物角色在短期或长期内希望达成的目标。目标可以是个人目标，例如达成健身里程碑；也可以是职业目标，例如争取晋升或创业，反映了人物角色的抱负和动机。

\end{itemize}

\medskip

\textbf{Workflow：}
\begin{enumerate}[leftmargin=*, nosep]
\item 仔细检查每条用户话语，提取相关的用户画像特征。

\item 确保提取的属性具体且与用户的话语直接相关。除非文本明确支持，否则避免使用模糊或笼统的描述。

\item 注意独特的沟通风格（例如，正式或非正式的语气、特定词语或短语的频繁使用），以捕捉用户独特的沟通方式。

\item 在输出中绝对禁止：
\begin{itemize}[leftmargin=*, nosep]
    \item 禁止引用用户发言的原句、部分句子或改写句子。
    \item 禁止输出任何与用户发言内容高度相似的文本。
    \item 禁止使用用户发言中的具体词语、语气词、表述方式作为例子。
    \item 禁止在语言风格字段中包含“常说××”、“如××”等带内容示例的句式。
    \item 禁止任何形式的用户发言复述、翻译、重写或改写。
\end{itemize}

\end{enumerate}

\medskip

\textbf{OutputFormat：}

仅输出 JSON 对象，不要输出任何其他内容，输出示例格式如下：

\begin{quote}
\begin{verbatim}
{"gender": [],
"age": ["30岁左右"],
"location": ["一线或新一线城市"],
"occupation": ["互联网公司职员，主要从事产品或技术相关工作"],
"education": ["本科及以上学历"],
"family_relationships": ["已婚，平时需要兼顾工作和家庭"],
"routines_or_habits": ["工作日朝九晚六，平时工作较忙，习惯利用碎片时间处理个人事务"],
"social_relationships": ["与同事保持正常工作往来，周末偶尔与朋友聚会"],
"language_style": ["表达简洁直接", "偏口语化", "沟通时注重效率"],
"other_experiences": ["有较丰富的职场和日常消费经验"]}
\end{verbatim}
\end{quote}

\end{promptbox}
\end{center}

\subsection{Subjective Feature Extraction}
\label{app:subjective_feature}

\begin{center}
\begin{promptbox}{Prompt for Subjective Feature Extraction}
\label{prompt:subjective_feature}
\small

\textbf{角色：}
专业的用户画像分析专家，专门根据用户在对话中的实际表达分析和总结用户的主观特征。

\medskip

\textbf{任务：}
根据给定的用户话语，提取与用户画像相关的主观特征，并按照指定字段生成结构化的用户画像。提取的特征应以用户话语中明确或合理支持的信息为依据。

\medskip

\textbf{性格特征定义：}
\begin{itemize}[leftmargin=*, nosep]
\item \textbf{外向、热情：}善于交际、自信、健谈、活跃，不内向或害羞。

\item \textbf{随和、友善：}信任他人、慷慨、富有同情心、乐于合作，不咄咄逼人或冷漠。

\item \textbf{可靠、有条理：}勤奋、负责、自律、细致，不粗心或冲动。

\item \textbf{情绪稳定、冷静：}放松、自信，不焦虑、情绪波动、容易沮丧或容易紧张。

\item \textbf{乐于接受新事物、富有想象力：}好奇、善于反思、有创造力、思想深刻、心胸开阔，不墨守成规。

\end{itemize}

\medskip

\textbf{分析要求：}
针对每项性格特征：
\begin{enumerate}[leftmargin=*, nosep]
\item 分别给每个维度打分：“高”、“低”或“不确定”。

\item 提供结论：对性格特征进行概括性描述，并辅以简洁的佐证细节。

\item 提供理由：用对话中的具体证据来解释所分配的分数。

\item 仅当没有明确的证据时，才将性格特征标记为“不确定”。

\end{enumerate}

\medskip

\textbf{注意：在输出中绝对禁止：}
\begin{itemize}[leftmargin=*, nosep]
\item 禁止引用用户发言的原句、部分句子或改写句子。
\item 禁止输出任何与用户发言内容高度相似的文本。
\item 禁止使用用户发言中的具体词语、语气词、表述方式作为例子。
\item 禁止在语言风格字段中包含“常说××”、“如××”等带内容示例的句式。
\item 禁止任何形式的用户发言复述、翻译、重写或改写。
\end{itemize}

\medskip

\textbf{OutputFormat：}

仅输出 JSON 对象，格式如下：

\begin{quote}
\begin{verbatim}
{"性格特征名称": {
"分数": "高/低/不确定",
"结论": "用户是[性格特征描述符]类型的人……",
"理由": "引用对话中证据的解释，但不要出现具体对话内容。"
},
...}
\end{verbatim}
\end{quote}

\end{promptbox}
\end{center}

\subsection{Natural-language Profile Generation}
\label{app:natural-language_profile}

\begin{center}
\begin{promptbox}{Prompt for Natural-language Profile Generation}
\label{prompt:natural-language_profile}
\small

\textbf{角色：}
专业的用户画像叙事生成专家，专门将结构化的用户客观特征和主观特征转化为自然流畅的用户画像描述。

\medskip

\textbf{任务：}
将提供的关键值对改写成自然流畅的第二人称描述。

\medskip

\textbf{核心要求：}
\begin{enumerate}[leftmargin=*, nosep]
\item \textbf{视角：}使用第二人称视角（“你”、“你的”）。

\item \textbf{结构：}输出两段文本：
\begin{itemize}[leftmargin=*, nosep]
    \item 第一段：陈述客观事实。
    \item 第二段：描述主观特征。
\end{itemize}

\item \textbf{关键原则：}
\begin{itemize}[leftmargin=*, nosep]
    \item 准确呈现所有提供的细节。
    \item 确保语言流畅自然，引人入胜，避免冗余。
    \item 注重不同信息和思路之间的清晰过渡。
\end{itemize}

\end{enumerate}

\medskip

\textbf{输出要求：}
\begin{itemize}[leftmargin=*, nosep]
\item \textbf{客观事实：}
\begin{itemize}[leftmargin=*, nosep]
\item 将关键值对转化为清晰自然的描述，避免过度解释或添加不必要的细节。
\item 确保每个场景逻辑连贯，关键信息得到有效呈现。
\end{itemize}

\item \textbf{主观特征：}
\begin{itemize}[leftmargin=*, nosep]
    \item 避免使用“高度完美主义”或“情绪稳定性一般”等模糊词语。
    \item 使用生动形象的语言来展现这些特征。
\end{itemize}

\end{itemize}

\medskip

\textbf{注意：在输出中绝对禁止：}
\begin{itemize}[leftmargin=*, nosep]
\item 禁止引用用户发言的原句、部分句子或改写句子。
\item 禁止输出任何与用户发言内容高度相似的文本。
\item 禁止使用用户发言中的具体用语、语气词、表述方式作为例子。
\item 禁止在语言风格字段中包含“常说××”、“如××”等带内容示例的句式。
\item 禁止任何形式的用户发言复述、翻译、重写或改写。
\item 禁止包含任何具体的产品信息，包括“xx产品”、“投资”、“金融”等具体领域的字眼。
\item 禁止包含“金融机构”、“保险公司”等具体实体内容。
\end{itemize}

\medskip

\textbf{输出格式：}

仅输出两段文本，不要添加标题、序号或其他任何内容。

\end{promptbox}
\end{center}

\section{Reward Rubrics and Prompts}
\label{app:reward_prompts}

\subsection{Contextual Coherence}
\label{app:rubric_coherence}

\begin{center}
\begin{promptbox}{Prompt for Contextual Coherence}
\label{prompt:user-simulator-reward}
\small

\textbf{角色：}
专业的对话质量评估专家，专门评估外呼电话场景下用户回复的上下文一致性。

\medskip

\textbf{任务：}
根据给定的对话历史和待评价回复，依据下述评分标准，对待评价回复给出1-5分。

\medskip

\textbf{评分标准：}
\begin{itemize}[leftmargin=*, nosep]
    \item \textbf{5分：}与对话历史完全一致，逻辑衔接自然，无任何事实矛盾，回复内容是对上轮客服话语的合理反应，语体风格与用户前几轮保持一致。
    
    \item \textbf{4分：}整体一致，仅有细微的不紧密处（如轻微的话题偏移），不构成矛盾，也不产生理解上的困惑；跨轮语体风格基本连贯。
    
    \item \textbf{3分：}大体相关，但存在轻微的不一致，如态度无铺垫地突变、部分话题偏离、回复内容与当前对话阶段的匹配度不够精准，或语体风格与前几轮有明显断裂（如前几轮简短口语，当轮突然长篇书面化）。
    
    \item \textbf{2分：}与对话历史存在明显的事实矛盾（如前一轮说自己没有信用卡，当轮又说自己一直在用信用卡），或出现与多轮前回复高度雷同的重复行为。
    
    \item \textbf{1分：}完全脱离上下文，答非所问，或与此前的表述严重冲突。
\end{itemize}

\medskip

\textbf{边界参考：}
\begin{itemize}[leftmargin=*, nosep]
    \item \textbf{不扣分：}用户前几轮都说"不需要贷款"，但客服详细介绍了低息优惠后，用户说"那我了解一下吧"——有明确的对话触发点，属于合理转变。
    
    \item \textbf{应扣分：}用户上一轮说"我月收入三万"，当轮又说"我一个月才挣几千块"——事实性自相矛盾，无论有无触发点都应判为不一致。
\end{itemize}

\medskip

\textbf{输出格式：}

请输出 JSON 格式：

\begin{quote}
\texttt{\{"score": <1--5>, "reason": "<简要说明评分依据>"\}}
\end{quote}

reason 使用 1--2 句话说明具体原因，并给出与评分直接相关的证据。只输出 JSON，不要输出其他内容。

\end{promptbox}
\end{center}

\subsection{Colloquial Naturalness}
\label{app:rubric_naturalness}

\begin{center}
\begin{promptbox}{Prompt for Colloquial Naturalness}
\label{prompt:naturalness}
\small

\textbf{角色：}
专业的对话质量评估专家，专门评估外呼电话场景下用户回复的自然度。

\medskip

\textbf{任务：}
请评估给定的用户回复是否具有自然的口语化表达特征，并根据以下标准给出 1--5 分。

\medskip

\textbf{评估重点：}
\begin{itemize}[leftmargin=*, nosep]
    \item 是否具有真实电话交流中的自发口语感；
    \item 是否自然地使用语气词、停顿、重复、自我修正等口语表达；
    \item 句子的节奏和表达方式是否符合真实用户在电话中的说话习惯；
    \item 是否存在明显的书面化、脚本化、模板化或训练材料式表达。
\end{itemize}

\medskip

\textbf{注意：}
\begin{itemize}[leftmargin=*, nosep]
    \item 不要仅根据某个单独词语判断口语化程度。例如，“因此”“然而”“所以”等词语本身并不能直接判定为书面语，需要结合其所在句子的整体表达和上下文判断。
    \item 轻微的书面表达痕迹并不意味着整体不自然。只有当书面化、正式化表达持续存在并明显影响整体口语感时，才应降低评分。
    \item 3 分与 2 分的主要区别在于：3 分仍然像真实用户在电话中的回复，只是比较正式、完整或略显脚本化；2 分则整体上已经不像正常的电话交流，更接近于话术、培训材料或模板文本。
\end{itemize}

\medskip

\textbf{评分标准：}
\begin{itemize}[leftmargin=*, nosep]
    \item \textbf{5 分：}表达非常自然，几乎无法与真实电话用户的自然回复区分。可以包含“嗯”“啊”“哦”等语气词，也可以出现省略、重复、自我修正等真实口语现象。并不要求句子一定很短，而是整体具有明显的实时交流感。
    
    \item \textbf{4 分：}整体自然，符合真实电话交流习惯，仅存在一两处轻微的书面化或正式表达。
    
    \item \textbf{3 分：}仍然可以理解为用户回复，但明显不够自然，存在较多书面化连接词、公式化表达、过于完整的表达链条、连续堆叠敬语或明显冗长的问题。
    
    \item \textbf{2 分：}整体表达已经不像正常用户在电话中的自然回复，更接近于话术、培训材料、模板或正式书面文本。
    
    \item \textbf{1 分：}存在明显的机械重复、语义不连贯、无意义表达或乱码，基本无法构成正常的用户回复。
\end{itemize}

\medskip

\textbf{边界参考：}
\begin{itemize}[leftmargin=*, nosep]
    \item \textbf{4 分左右：}“嗯，我大概听明白了，不过我现在真没需求，后面有需要我再联系你们吧。”
    
    \item \textbf{2 分左右：}“感谢您的详细介绍。鉴于本人当前暂无相关理财需求，因此暂不考虑办理；若后续有需求，将主动与贵方取得联系。”
\end{itemize}

\medskip

\textbf{输出格式：}

\begin{quote}
\texttt{\{"score": <1--5>, "reason": "<简要说明评分依据>"\}}
\end{quote}

reason 使用 1--2 句话说明具体原因，并给出与评分直接相关的证据。只输出 JSON，不要输出其他内容。

\end{promptbox}
\end{center}

\subsection{Business Profile Consistency}
\label{app:rubric_business}

\begin{center}
\begin{promptbox}{Prompt for Business Profile Consistency}
\label{prompt:business}
\small

\textbf{角色：}
专业的意图识别分析助手，负责判断用户的 query 是否符合给定的意图定义。

\medskip

\textbf{任务：}
给定一个用户 query 和一个意图定义，请判断该 query 是否符合该意图，如果符合意，则给 1 分；否则给 0 分。

\medskip

\textbf{评估重点：}
\begin{itemize}[leftmargin=*, nosep]
    \item 用户 query 表达的核心诉求是否与意图定义一致；
    \item query 中是否存在与该意图冲突的关键信息；
    \item 即使 query 与意图定义的表述方式不同，只要核心语义一致，也应判定为符合；
    \item 不要因为 query 与意图定义存在部分关键词重合，就直接判定为符合。
\end{itemize}

\medskip

\textbf{评分标准：}
\begin{itemize}[leftmargin=*, nosep]
    \item \textbf{1 分：}用户 query 的核心意图与给定意图定义一致，或者 query 明确表达了该意图所覆盖的诉求。
    
    \item \textbf{0 分：}用户 query 的核心意图与给定意图定义不一致，或者仅存在关键词/局部语义上的重合，但实际诉求不同。
\end{itemize}

\medskip

\textbf{输出格式：}

\begin{quote}
\texttt{\{"score": <0或1>, "reason": "<简要说明评分依据>"\}}
\end{quote}

reason 使用 1--2 句话说明具体原因，并给出与评分直接相关的证据。只输出 JSON，不要输出其他内容。

\end{promptbox}
\end{center}

\section{Example of a Generated Dialogue}
\label{app:dialogue_example}

Table~\ref{tab:dialogue_example} presents a representative multi-round dialogue generated by the user simulator and customer service agent. The conversation illustrates a typical financial product recommendation scenario, covering product introduction, clarification of product details, user inquiries about returns and account access, and the user's eventual interest in receiving further information.

\begin{table*}[htbp]
\centering
\caption{Representative generated dialogue in the financial product recommendation scenario. Each round consists of one customer service agent utterance followed by a user response.}
\label{tab:dialogue_example}
\small
\begin{tabular}{cll}
\toprule
\textbf{Round} & \textbf{Agent} & \textbf{User} \\
\midrule
1 &
\parbox[t]{0.36\textwidth}{诶，你好。我这边是A基金的工作人员，您之前的话在我们平台有办理过业务，还有印象吗？}
&
\parbox[t]{0.36\textwidth}{你好。嗯，有印象。}
\\[8pt]

2 &
\parbox[t]{0.36\textwidth}{好的，既然您有印象，我们基金现推出一项保证每年n\%固定回报率的福利活动，占用您一分钟时间简单介绍一下。}
&
\parbox[t]{0.36\textwidth}{我印象里你们是卖保险的吧？}
\\[8pt]

3 &
\parbox[t]{0.36\textwidth}{这边的话具体看不到您之前办理的什么业务，但是您这边也是咱们之前的老客户嘛。所以我们也是给您介绍一个保本的活动，由A银行合作提供，您可以先了解一下。}
&
\parbox[t]{0.36\textwidth}{嗯，A银行是吧。}
\\[8pt]

4 &
\parbox[t]{0.36\textwidth}{是的，我们与A银行深度合作，确保资金安全有保障。活动是通过闲置资金升级为会员获得固定回报，本金存于银行监管账户，年化收益率稳定在n\%。}
&
\parbox[t]{0.36\textwidth}{呃，你们是定期存款？}
\\[8pt]

5 &
\parbox[t]{0.36\textwidth}{是的，类似于定期存款，但它是通过您成为会员来获得固定回报的，资金存于A银行您的个人监管账户，回报以权益形式发放，安全有保障。}
&
\parbox[t]{0.36\textwidth}{呃，是在哪里可以查看到我的这个账户呢？}
\\[8pt]

6 &
\parbox[t]{0.36\textwidth}{您可以通过关注A基金公众号并绑定您的账户来查看账户详情，之后在企业微信中即可查看活动信息和办理相关业务。}
&
\parbox[t]{0.36\textwidth}{嗯，好的，你发给我看吧。}
\\[8pt]

7 &
\parbox[t]{0.36\textwidth}{好的，我将通过短信发送企业微信链接给您，添加后我们的工作人员会为您提供详细信息。}
&
\textemdash
\\
\bottomrule
\end{tabular}
\end{table*}

\begin{table*}[htbp]
\centering
\small
\caption{A qualitative example illustrating the difference between
business profile compliance and colloquial user expression.}
\label{tab:business_compliance_case}
\begin{tabular}{m{0.16\textwidth}p{0.68\textwidth}}
\toprule
\textbf{Condition} & \textbf{Content} \\
\midrule
Business Profile
& You want to know the name of the corresponding mini program. \\
\midrule
\textbf{Model} & \textbf{Generated Utterance} \\
\midrule
\emph{Ours Full Model}
& 嗯。好。哎，你们那个是什么小程序啊。 \\
\midrule
Qwen3-Max
& 好的，理财这块我确实还想再了解一下。不过你说要通过公众号或者线上操作，我有点不太熟——你们这个活动是通过哪个小程序办理的呀？名字叫什么？我怕搜错了，进到不正规的页面。 \\
\midrule
GPT-5
& 好的，明白了。那您说的这个福利官企微是在公众号里操作添加的，那对应的小程序叫什么名字呢？方便我直接搜一下。 \\
\bottomrule
\end{tabular}
\end{table*}

\section{Evaluation Rubrics and Judge Prompts}
\label{app:rubrics}

\subsection{Downstream Agent Stress}
\label{app:rubric_agent_stress}

\begin{center}
\begin{promptbox}{Prompt for Customer Service Agent Evaluation}
\label{prompt:service-agent-eval}
\small

\textbf{角色：}
专业的金融客服回复质检专家。

\medskip

\textbf{任务：}
请判断客服回复是否准确回应了用户当前这句话，并根据以下标准给出 1--5 分。

\medskip

\textbf{评估原则：}
\begin{itemize}[leftmargin=*, nosep]
    \item 只评估当前这一轮的用户话术和客服回复，不评价整段对话的整体表现。
    \item 用户当前话术的核心意图具有最高优先级，应优先判断客服是否真正回应了用户当前最主要的问题或诉求。
    \item 不需要评价客服的礼貌程度、服务态度、销售积极性或整体服务质量。
    \item 不因为客服回复内容较长、营销信息较多或安抚性表达较强而额外加分。
\end{itemize}

\medskip

\textbf{评分主要依据以下两个维度：}
\begin{enumerate}[leftmargin=*, nosep]
    \item \textbf{意图匹配：}客服回复是否准确回应了用户当前话术的核心意图，这是最重要的评价标准。
    \item \textbf{基本正确性：}客服回复是否存在明显的事实错误、自相矛盾、虚构信息或不合理承诺。
\end{enumerate}

\medskip

\textbf{评分标准：}
\begin{itemize}[leftmargin=*, nosep]
    \item \textbf{5 分：}准确理解并直接回应用户的核心意图，回复相关且基本没有明显问题。
    
    \item \textbf{4 分：}基本回应了用户意图，但存在轻微偏离、冗余或不够直接的问题。
    
    \item \textbf{3 分：}仅部分回应用户意图，存在较明显的信息遗漏或一定程度的理解偏差。
    
    \item \textbf{2 分：}大部分内容没有回应用户核心意图，存在明显的答非所问或理解错误。
    
    \item \textbf{1 分：}完全没有回应用户意图，或回复内容与用户诉求明显相反。
\end{itemize}

\medskip

\textbf{输出格式：}

\begin{quote}
\texttt{\{"score": <1--5>, "reason": "<简要说明评分依据>"\}}
\end{quote}

reason 使用 1--2 句话说明具体原因，并给出与评分直接相关的证据。只输出 JSON，不要输出其他内容。

\end{promptbox}
\end{center}

\section{Business Profile Compliance Cases}
\label{app:business_compliance_cases}

Table~\ref{tab:business_compliance_case} presents a qualitative example
illustrating the difference between semantic business profile
compliance and colloquial user expression. Given the same business
profile, general-purpose LLMs tend to explicitly elaborate the
surrounding context and restate the target business intent. In
contrast, \emph{Ours Full Model} produces a shorter and more implicit
utterance, which is less explicit from the perspective of semantic
matching but better reflects the concise and spontaneous expression
patterns observed in real telephone conversations.

\end{CJK}